\documentclass[10pt,twocolumn,letterpaper]{article}

\usepackage{cvpr}              

\usepackage{graphicx}
\usepackage{amsmath}
\usepackage{amssymb}
\usepackage{booktabs}
\usepackage[super]{nth}

\usepackage[pagebackref,breaklinks,colorlinks]{hyperref}

\usepackage[capitalize]{cleveref}
\crefname{section}{Sec.}{Secs.}
\Crefname{section}{Section}{Sections}
\Crefname{table}{Table}{Tables}
\crefname{table}{Tab.}{Tabs.}

\def\paperID{32} 
\def\confName{CVPR}
\def\confYear{2024}

\begin{document}

\title{Spatially Optimized  Compact Deep Metric Learning Model for Similarity Search}

\author{Md. Farhadul Islam$^1$, Md. Tanzim Reza$^1$, Meem Arafat Manab$^{1, 2}$, \\
Mohammad Rakibul Hasan Mahin$^1$, Sarah Zabeen$^1$, Jannatun Noor$^1$\\
$^1$School of Data and Sciences, BRAC University, Dhaka, Bangladesh\\
$^2$School of Law and Government, Dublin City University, Dublin, Ireland\\
{\tt\small \{farhadul.islam, tanzim.reza, meem.arafat, ext.mohammad.rakibul.hasan,} \\ {\tt\small sarah.zabeen, jannatun.noor\}@bracu.ac.bd}}

\maketitle

\begin{abstract}
   Spatial optimization is often overlooked in many computer vision tasks. Filters should be able to recognize the features of an object regardless of where it is in the image. Similarity search is a crucial task where spatial features decide an important output. The capacity of convolution to capture visual patterns across various locations is limited. In contrast to convolution, the involution kernel is dynamically created at each pixel based on the pixel value and parameters that have been learned. This study demonstrates that utilizing a single layer of involution feature extractor alongside a compact convolution model significantly enhances the performance of similarity search. Additionally, we improve predictions by using the GELU activation function rather than the ReLU. The negligible amount of weight parameters in involution with a compact model with better performance makes the model very useful in real-world implementations. Our proposed model is below 1 megabyte in size. We have experimented with our proposed methodology and other models on CIFAR-10, FashionMNIST, and MNIST datasets. Our proposed method outperforms across all three datasets.
\end{abstract}

\section{Introduction}
\label{sec:intro}
Metric learning is the process of calculating the similarity or dissimilarity between pairs or triplets of objects \cite{bellet2022metric}, which has been put to application in various use cases such as face recognition \cite{schroff2015facenet}, Natural Language Processing \cite{wohlwend2019metric}, recommendation systems \cite{cao2007learning}, healthcare \cite{zhong2021deep}, and so on. Although simple distance calculation metrics such as Euclidean distance may serve the purpose hypothetically, in reality, such a standard approach may ignore domain-specific important features of data \cite{suarez2021tutorial}. Thus, different types of advanced distance metrics are used for distance calculation \cite{moutafis2016overview}. Regardless of distance metrics, the majority of the metric calculation models are dominated by convolution architectures, where a convolution-based model extracts learned feature embeddings, and distance metric is calculated based on it. However, convolution operation has its share of limitations due to limited local receptive field, making it unable to retrieve information from global spatial relations. Involution addresses this challenge by employing a dynamic kernel while remaining lightweight. Moreover, in applications where spatial context is necessary, involution often performs well even as an addition to convolution\cite{zhang2021automatic, shao2022spatial, jain2023coinnet, islam2024involution, 9665787, 10115109}. Hence, we propose a similarity-finding model based on involution-convolution, designed to handle global spatial relations efficiently while maintaining low resource usage. 

For the proposed metric learning model, we employed the Categorical Cross-Entropy (CE) and Multi-Similarity (MS) \cite{wang2019multi} loss. As CE can be prohibitive in case of large class size and lack of absolute labels \cite{10.1007/978-3-030-58539-6_33}, additional experiments with pair-wise MS loss should complement it and provide further context. Our research primarily contributes by introducing a spatially optimized compact Deep Metric Model incorporating involution and GELU activation. Moreover, we conduct empirical analysis using two loss functions, MS and CE, across three general datasets to evaluate performance. Our empirical analysis showcases that the single involution layer-based hybrid model performs quite competitively against traditional convolution-based architectures while being extremely lightweight. Our proposed method is simple to implement yet effective unlike other hybrid models of involution and convolution.

\section{Background}
\label{sec:back}

\subsection{Involution}

Involution \cite{Li_2021_CVPR} dynamically generates a kernel for each pixel. This dynamic creation relies on the pixel's individual numerical value and parameters of learning. This approach proves beneficial as it resembles attention to input, thereby regulating the weights of the kernel. Involution computes every output pixel by applying Equation \ref{eq1}:.


\begin{equation}
\label{eq1}
Y_{i,k,k} = \sum_{(u,u)\in \Delta_{K} }^{} H_{i,j,u + \lfloor K/2 \rfloor , v + \lfloor K/2 \rfloor , \lceil kG/C \rceil } X_{i+u,j+v,k}
\end{equation}

In the channel dimension, groups of channels share the same kernel. The kernel is produced using the linear network shown below in Equation \ref{eq2}.

\begin{equation}
\label{eq2}
 H_{i,j} = \phi(X_{i,j}) = W_{1} \sigma(W_{0}X_{i,j})
\end{equation}

The meta weights ($W_1$, $W_0$) used to construct the kernel are shared across all pixels, preserving convolution's shift-invariance to some extent. While involution doesn't capture interpixel interactions as effectively as attention, its linear complexity compensates for it sufficiently. As a result, there's no need to track different kernel weights; only the meta weights are necessary. This simplification enables the creation of more complex models than with convolution, reducing the number of weight parameters even with multiple layers of involution. However, the multiple additions of involution layers increase the training time required.

\subsection{Gaussian Error Linear Units}

The activation function known as Gaussian Error Linear Unit (GELU) \cite{hendrycks2023gaussian} is denoted by $x\Phi(x)$ and the standard Gaussian cumulative distribution function is denoted by ($\Phi(x)$). Unlike Rectified Linear Units (ReLU), which determine input gating based on their sign ($x\mathbf{1}_{x>0}$), GELU assigns weights to inputs based on their percentile, offering a nuanced way to handle input values. This methodology results in a smoother nonlinearity compared to ReLUs, thereby introducing a more gradual transition between activation states. In essence, GELU serves as a refined alternative to ReLUs, providing enhanced flexibility and performance in various neural network architectures.
$$\text{GELU}\left(x\right) = x{P}\left(X\leq{x}\right) = x\Phi\left(x\right) = x \cdot \frac{1}{2}\left[1 + \text{erf}(x/\sqrt{2})\right],$$
if $X\sim \mathcal{N}(0,1)$

\section{Proposed Methodology}

As mentioned in \cite{9665787}, a combination of involution and convolution results in enhanced feature extraction. Moreover, it lessens the model size due to less use of convolutions. Our case is similar and proposes a similar yet more effective approach. Involution and convolution can be represented as \ref{3_eq} and \ref{4_eq}. Here, $X$ represents the input data, $F$ denotes the involution kernel, and $\sigma$ encompasses a non-linear activation function (GELU), involving data normalization. $\mathcal{H}$ signifies the convolution kernel, while $\delta$ denotes the activation function (GELU).

\begin{figure}[ht!]
    \makebox[\linewidth]{
        \includegraphics[width=0.90\linewidth]{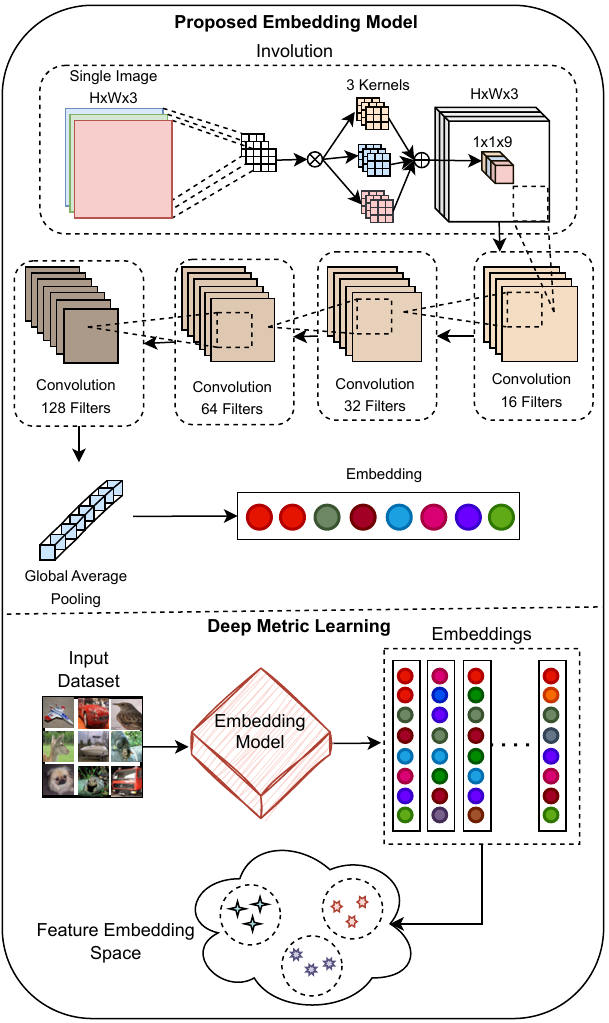}}

    \caption{Proposed methodology of the metric learning network. The proposed hybrid network begins with a single layer of involution.}
    \label{fig:proposed}
\end{figure}

\begin{equation} 
y_{inv}=\sigma (F \ast X)
\label{3_eq}
\end{equation}

\begin{equation} 
y_{conv}=\delta (\mathcal {H} \ast y_{inv})
\label{4_eq}
\end{equation}

To fully utilize the feature representation in the channel domain, the output shape would be automatically aligned to the input feature in the spatial dimension. Convolution, on the other hand, is used to enhance the possible variety of feature representation by complementing the channel-agnostic in involution. Our proposed model consists of a single involution layer which takes the initial input, followed by 4 convolution layers. The involution layer receives the 3-channel image and produces a feature map of the same dimension, which should have a more specialized feature map compared to general convolution due to its global receptive natures. The subsequent convolution layers have 16, 32, 64, and 128 filters, respectively, followed by a global average pooling layer converting the feature volume to a single-dimensional one. Finally, a single fully connected layer creates the embedding which can be used to find similarities/differences against other embeddings. CE and MS loss function-driven training is used to minimize the distance of embedding from the same classes and maximize the distance from different classes. Our model uses Gaussian Error Linear Unit as activation functions between layers instead of Rectified Linear Unit, as GeLU is differentiable at every point, unlike rectifiers, and can retain image distance metrics more successfully after applying \cite{lee2023gelu, 9538437}. ReLU is often preferred over other activation functions due to its extremely low computational cost and has been lately surpassed by GeLU and other nonlinear functions such as Sigmoid Linear Units to, among other things, resolve the vanishing gradient problem \cite{DUBEY202292, JMLR:v25:23-0912}. Our task, however, specifically calls for Gaussian Error, as Gaussian Errors closely mirror the visual cues associated with distance metrics among images. Our proposed methodology is demonstrated in Figure \ref{fig:proposed}.

\section{Experiments and Results}
\label{sec:res}

This study uses an NVIDIA GeForce RTX 3080Ti GPU, which has a performance of 34.1 TeraFLOPS, to train and assess the models. For empirical comparisons, we utilize MNIST, FashionMNIST, and CIFAR-10 datasets. We use CE loss \cite{10.1007/978-3-030-58539-6_33} and MS loss \cite{wang2019multi} for getting two perspectives of evaluation as mentioned earlier. 

There are two vanilla convolutional neural network (CNN) variants with three convolution layers, `3a' (\nth{2} layer having 64 filters) and `3b' (\nth{2} layer having 96 filters). There are in total three vanilla involutional neural network (INN) variants. Vanilla INN-2 has 2 involution layers and vanilla INN-3 and INN-4 have 3 and 4 involution layers respectively. Hybrid models in this experiment have 3 convolution layers with 64 nodes in the second layer like vanilla CNN-3a. Three variations have 1 to 3 layers of involutions and are addressed as hybrid-1, hybrid-2, and hybrid-3. To avoid slow training involutions are used before convolutions. Both operations have the same stride of $1\times1$ size. For comparison, we have used MobileNetV2, EfficientNetV2B0, ResNet50V2, VGG16, and NASNetMobile considering their size and training time per epoch. The learning rate is 0.001 with the Adam optimizer and all models are trained for 20 epochs in MNIST and FashionMNIST and 25 epochs in CIFAR-10.


\begin{figure}[ht!]
          \centering
          
          \begin{subfigure}[b]{0.47\textwidth}
            \includegraphics[width=\textwidth]{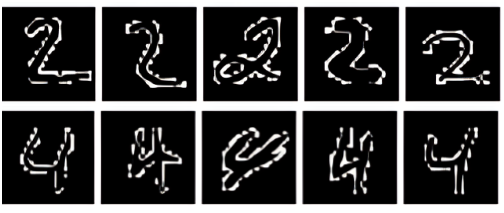}
            \centering
            \caption{Predictions from MNIST}
            \label{mnist:similarity}
          \end{subfigure}          
          
          \begin{subfigure}[b]{0.47\textwidth}
            \includegraphics[width=\textwidth]{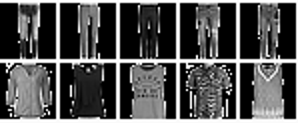}
            \centering
            \caption{Predictions from FashionMNIST}
            \label{f-mnist:similarity}
          \end{subfigure}
          
          \begin{subfigure}[b]{0.47\textwidth}
            \includegraphics[width=\textwidth]{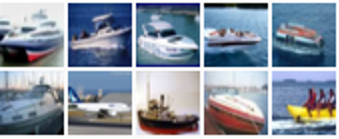}
            \centering
            \caption{Predictions from CIFAR-10}
            \label{cifar:similarity}
          \end{subfigure}

            \caption{Sample predictions of similarity search.}
        \label{fig:simi}
        \end{figure}

\begin{table*}[ht!]
\centering
\caption{Comparison among similar variations of the proposed model on MNIST, FashionMNIST, and CIFAR-10 datasets. Here, $i$ and $c$ represent the number of involution and convolution layers respectively. CrossEntropy is shortened as `CE' and Multi-Similarity is shortened as `MS'.}
\resizebox{\textwidth}{!}{%
\begin{tabular}{c|c|c|cccc|c|c|cc}
\hline
{Model} & {\begin{tabular}[c]{@{}c@{}}Weight \\ Parameters\end{tabular}} & {\begin{tabular}[c]{@{}c@{}}Model \\ Size\end{tabular}} & \multicolumn{4}{c|}{Image Size = $28 \times 28$} & {\begin{tabular}[c]{@{}c@{}}Weight \\ Parameters\end{tabular}} & {\begin{tabular}[c]{@{}c@{}}Model \\ Size\end{tabular}} & \multicolumn{2}{c}{\begin{tabular}[c]{@{}c@{}}Image Size = \\ $32 \times 32$\end{tabular}} \\ \cline{4-7} \cline{10-11} 
 &  &  & \multicolumn{2}{c|}{MNIST} & \multicolumn{2}{c|}{FashionMNIST} &  &  & \multicolumn{2}{c}{CIFAR-10} \\ \cline{4-7} \cline{10-11} 
 &  &  & \multicolumn{1}{c|}{CE Loss} & \multicolumn{1}{c|}{MS Loss} & \multicolumn{1}{c|}{CE Loss} & MS Loss &  &  & \multicolumn{1}{c|}{CE Loss} & MS Loss \\ \hline
Vanilla CNN-3a & 116,320 & 454.38 KB & \multicolumn{1}{c|}{0.0757} & \multicolumn{1}{c|}{0.0679} & \multicolumn{1}{c|}{0.441} & 0.6214 & 116,608 & 455.50 KB & \multicolumn{1}{c|}{1.636} & 1.116 \\
Vanilla CNN-3b & 157,824 & 616.50 KB & \multicolumn{1}{c|}{0.0756} & \multicolumn{1}{c|}{0.0839} & \multicolumn{1}{c|}{0.436} & 0.6078 & 158,112 & 617.62 KB & \multicolumn{1}{c|}{1.426} & 1.189 \\
Vanilla INN-2 & 560 & 2.19 KB & \multicolumn{1}{c|}{1.8941} & \multicolumn{1}{c|}{1.1249} & \multicolumn{1}{c|}{2.134} & 1.2155 & 1,076 & 4.20 KB & \multicolumn{1}{c|}{1.575} & 1.233 \\
Vanilla INN-3 & 584 & 2.38 KB & \multicolumn{1}{c|}{1.8747} & \multicolumn{1}{c|}{1.1349} & \multicolumn{1}{c|}{1.925} & 1.1641 & 1,102 & 4.30 KB & \multicolumn{1}{c|}{1.580} & 1.233 \\
Vanilla INN-4 & 608 & 2.38 KB & \multicolumn{1}{c|}{1.9023} & \multicolumn{1}{c|}{1.1338} & \multicolumn{1}{c|}{1.713} & 1.2024 & 1,158 & 4.40 KB & \multicolumn{1}{c|}{1.606} & 1.334 \\
Hybrid-2 ($i=2, c=3$) & 116,368 & 454.56 KB & \multicolumn{1}{c|}{0.0801} & \multicolumn{1}{c|}{0.1191} & \multicolumn{1}{c|}{0.445} & 0.6401 & 116,660 & 455.70 KB & \multicolumn{1}{c|}{1.333} & 1.007 \\
Hybrid-3 ($i=3, c=3$)  & 116,392 & 454.66 KB & \multicolumn{1}{c|}{0.085} & \multicolumn{1}{c|}{0.1259} & \multicolumn{1}{c|}{0.454} & 0.6212 & 116,686 & 455.80 KB & \multicolumn{1}{c|}{1.401} & 1.16 \\
\textbf{Hybrid-1} ($i=1, c=3$)  & 116,344 & 454.47 KB & \multicolumn{1}{c|}{0.749} & \multicolumn{1}{c|}{0.0707} & \multicolumn{1}{c|}{0.433} & 0.6110 & 116,634 & 455.60 KB & \multicolumn{1}{c|}{1.309} & 0.932 \\ \hline
\end{tabular}%
}
\label{boro_tab}
\end{table*}

\begin{table}[ht!]
\centering
\caption{Comparison with other efficient convolution-based models on $32\times32$ CIFAR-10 dataset. CrossEntropy is shortened as `CE', and Multi-Similarity is shortened as `MS'. The best performance is bolded and the second best is in italic font. `Params' or weight parameters are in millions and Size is in MegaBytes.}
\resizebox{\linewidth}{!}{%
\begin{tabular}{c|c|c|c|cc}
\hline
{Model} & {Params} & {Size} & {\begin{tabular}[c]{@{}c@{}} Seconds \\ Per Epoch\\ (Average)\end{tabular}} & \multicolumn{2}{c}{CIFAR-10} \\ \cline{5-6} 
 &  &  &  & \multicolumn{1}{c|}{CE Loss} & MS Loss \\ \hline
Vanilla CNN-3b & \textit{0.157} & \textit{0.62} & \textbf{5} & \multicolumn{1}{c|}{1.426} & 1.116 \\
EfficientNetB0 & 4 & 15.5 & 95  & \multicolumn{1}{c|}{1.703} & 1.569 \\
MobileNetV2 & 2.2 & 8.6 & 41  & \multicolumn{1}{c|}{1.623} & 1.429 \\
ResNet50V2 & 23 & 90 & 53  & \multicolumn{1}{c|}{\textbf{1.286}} & \textbf{0.901} \\
VGG16 & 14 & 56 & 16  & \multicolumn{1}{c|}{1.729} & 1.559 \\
NASNetMobile & 4.2 & 16 & 213  & \multicolumn{1}{c|}{1.357} & 1.023 \\
\textbf{Ours (Hybrid-1)} & \textbf{0.116} & \textbf{0.45} & \textit{7} & \multicolumn{1}{c|}{\textit{1.309}} & \textit{0.932} \\ \hline
\end{tabular}%
}
\label{choto_tab}
\end{table}



\begin{figure*}[ht!]
    \centering
    \begin{subfigure}[t]{0.50\textwidth}
        \centering
        \includegraphics[height=1.5in]{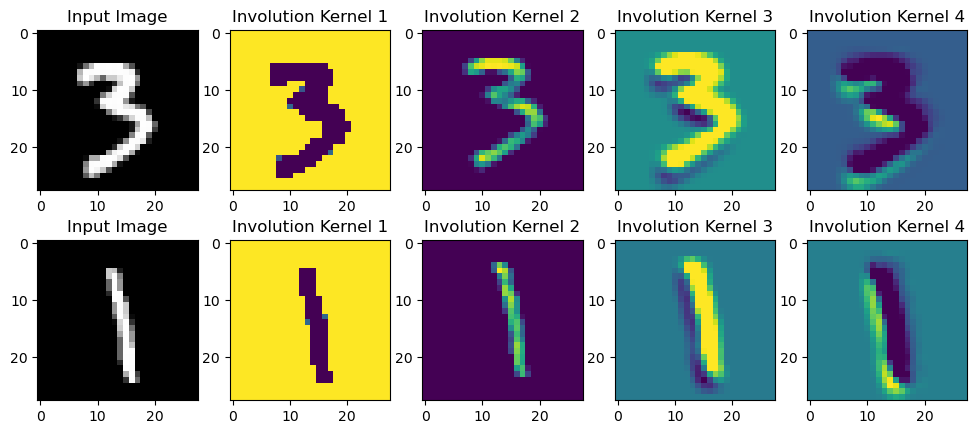}
        \caption{Involution kernel maps of MNIST.}
        \label{fig:pred1}
    \end{subfigure}%
    ~ 
    \begin{subfigure}[t]{0.50\textwidth}
        \centering
        \includegraphics[height=1.5in]{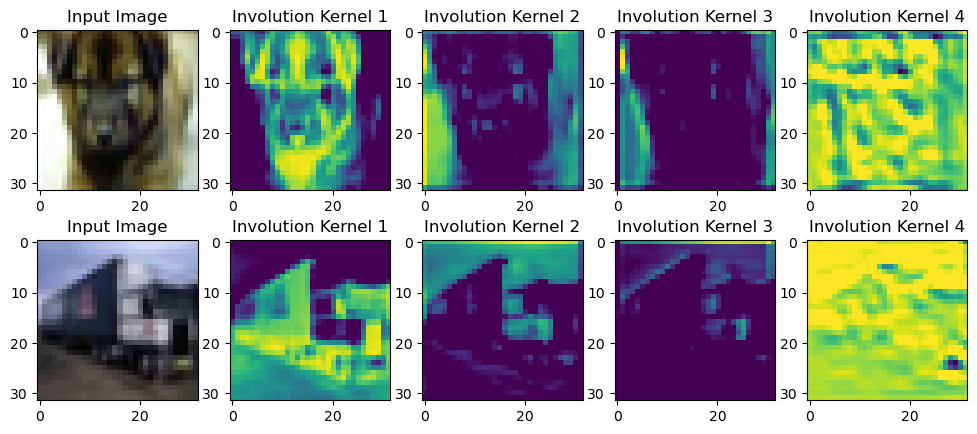}
        \caption{Involution kernel maps of CIFAR-10.}
        \label{fig:pred2}
    \end{subfigure}
    \caption{Involution kernel visualizations, showing how multiple involutions result in information loss and redundancy.}
    \label{fig:multiple}
\end{figure*}

In Table \ref{boro_tab} we find that our proposed hybrid-1 model outperforms all in terms of both cross-entropy loss and multi-similarity loss. Considering the performance of INN variations, the fewer weight parameters are not helpful with such poor results. The comparison with deeper models can be seen in Table \ref{choto_tab}. Only ResNet50V2 performs well here but with 23 Million weight parameters; ours performs similarly with around 100 thousand weight parameters. The reasonings will be discussed later in Section \ref{disc}. The similarity search predictions can be seen in Figure \ref{fig:simi} (Figure \ref{mnist:similarity} for MNIST, Figure \ref{f-mnist:similarity} for FashionMNIST, and \ref{cifar:similarity} for CIFAR-10).

\section{Discussion}
\label{disc}
Our empirical research unearths some of the more crucial aspects of metric learning. By using widely-used benchmark datasets, we inspected how the introduction of involution layers in an otherwise convolutional network can better performance measures while decreasing the model's size. For example, while a vanilla model of 3 layers of CNN performs fairly well (multi-similarity loss 1.116), the inclusion of one single layer of involution decreases this loss to 0.932, while also decreasing the size of the model by nearly 35\%, from 157,824 parameters to 116,344 parameters. Its performance, in terms of both cross-entropy and multi-similarity loss, is only rivaled by ResNet50v2, whose only feature extraction layers are its 48 convolution layers operating through skip connections. The other deeper models like EfficientNetB0 or VGG16, while shallower than ResNet and thus smaller, employ ingenious architectural structures that seem to have failed for learning pairwise metric distances, although they have stood the test of time for other more independent tasks like image classification or segmentation. Our model secures the second best performance among the ones tested, while also proving to be the smallest one among them. Among other measures, including training time, our proposed model consistently ranks among the top two.

Table \ref{boro_tab} also demonstrably shows that the insertion of one involution layer is enough for performance boosting, while the insertion of more than one instead decreases the performance. The root of this lies in Equation \ref{eq2}, which is used for constructing the involution kernel. The kernel captures location-specific features from the image, which, when applied more than once, leads to a loss of information in larger and more complicated datasets like CIFAR-10. CIFAR-10's images show higher diversity and hence greater distance metrics within the same group, so any clustering or metric extraction layer should leave some distance metrics unexplored for the next layers, but as Figure \ref{fig:multiple} proves, data is instead lost when three involution layers are successfully applied to its images. Compared to CIFAR-10, two or three layers of involution perform better for datasets MNIST, where in-group distance metrics are smaller and out-group metrics are considerably larger. Such data and their metric differences are not easily distorted when multiple layers of involution are executed.

\section{Conclusion and Future Work}

Our initial investigation into the application of involution layers in metric learning demonstrates that they can significantly enhance performance when integrated with convolutional architecture. As this also reduces the model's size in terms of parameters, this could have great implications for both transformer-based and CNN-based models. The extension of this particular research will focus on larger datasets. to further justify this proposed methodology's strength against transformer-based models. Some of the more exciting works that could potentially stem from here include using our model for low-cost metric learning as a component of information retrieval and taxonomy extraction and using metric learning for local feature matching in real-world locational and structural image datasets.

{\small
\bibliographystyle{ieee_fullname}
\bibliography{main}

\begin{thebibliography}{10}\itemsep=-1pt

\bibitem{bellet2022metric}
Aur{\'e}lien Bellet, Amaury Habrard, and Marc Sebban.
\newblock {\em Metric learning}.
\newblock Springer Nature, 2022.

\bibitem{10.1007/978-3-030-58539-6_33}
Malik Boudiaf, J{\'e}r{\^o}me Rony, Imtiaz~Masud Ziko, Eric Granger, Marco Pedersoli, Pablo Piantanida, and Ismail~Ben Ayed.
\newblock A unifying mutual information view of metric learning: Cross-entropy vs. pairwise losses.
\newblock In Andrea Vedaldi, Horst Bischof, Thomas Brox, and Jan-Michael Frahm, editors, {\em Computer Vision -- ECCV 2020}, pages 548--564, Cham, 2020. Springer International Publishing.

\bibitem{cao2007learning}
Zhe Cao, Tao Qin, Tie-Yan Liu, Ming-Feng Tsai, and Hang Li.
\newblock Learning to rank: from pairwise approach to listwise approach.
\newblock In {\em Proceedings of the 24th international conference on Machine learning}, pages 129--136, 2007.

\bibitem{DUBEY202292}
Shiv~Ram Dubey, Satish~Kumar Singh, and Bidyut~Baran Chaudhuri.
\newblock Activation functions in deep learning: A comprehensive survey and benchmark.
\newblock {\em Neurocomputing}, 503:92--108, 2022.

\bibitem{hendrycks2023gaussian}
Dan Hendrycks and Kevin Gimpel.
\newblock Gaussian error linear units (gelus), 2023.

\bibitem{islam2024involution}
Md.~Farhadul Islam, Meem~Arafat Manab, Joyanta~Jyoti Mondal, Sarah Zabeen, Fardin~Bin Rahman, Md.~Zahidul Hasan, Farig Sadeque, and Jannatun Noor.
\newblock Involution fused convnet for classifying eye-tracking patterns of children with autism spectrum disorder, 2024.

\bibitem{10115109}
Md.~Farhadul Islam, Sarah Zabeen, Fardin~Bin Rahman, Md.~Azharul Islam, Fahmid~Bin Kibria, Meem~Arafat Manab, Dewan~Ziaul Karim, and Annajiat~Alim Rasel.
\newblock Unic-net: Uncertainty aware involution-convolution hybrid network for two-level disease identification.
\newblock In {\em SoutheastCon 2023}, pages 305--312, 2023.

\bibitem{jain2023coinnet}
Samir Jain, Rohan Atale, Anubhav Gupta, Utkarsh Mishra, Ayan Seal, Aparajita Ojha, Joanna Kuncewicz, and Ondrej Krejcar.
\newblock Coinnet: A convolution-involution network with a novel statistical attention for automatic polyp segmentation.
\newblock {\em IEEE Transactions on Medical Imaging}, 2023.

\bibitem{lee2023gelu}
Minhyeok Lee.
\newblock Gelu activation function in deep learning: A comprehensive mathematical analysis and performance, 2023.

\bibitem{Li_2021_CVPR}
Duo Li, Jie Hu, Changhu Wang, Xiangtai Li, Qi She, Lei Zhu, Tong Zhang, and Qifeng Chen.
\newblock Involution: Inverting the inherence of convolution for visual recognition.
\newblock In {\em Proceedings of the IEEE/CVF Conference on Computer Vision and Pattern Recognition (CVPR)}, pages 12321--12330, June 2021.

\bibitem{9665787}
Guihuang Liang and Haoxiang Wang.
\newblock I-cnet: Leveraging involution and convolution for image classification.
\newblock {\em IEEE Access}, 10:2077--2082, 2022.

\bibitem{moutafis2016overview}
Panagiotis Moutafis, Mengjun Leng, and Ioannis~A Kakadiaris.
\newblock An overview and empirical comparison of distance metric learning methods.
\newblock {\em IEEE transactions on cybernetics}, 47(3):612--625, 2016.

\bibitem{9538437}
Anh Nguyen, Khoa Pham, Dat Ngo, Thanh Ngo, and Lam Pham.
\newblock An analysis of state-of-the-art activation functions for supervised deep neural network.
\newblock In {\em 2021 International Conference on System Science and Engineering (ICSSE)}, pages 215--220, 2021.

\bibitem{schroff2015facenet}
Florian Schroff, Dmitry Kalenichenko, and James Philbin.
\newblock Facenet: A unified embedding for face recognition and clustering.
\newblock In {\em Proceedings of the IEEE conference on computer vision and pattern recognition}, pages 815--823, 2015.

\bibitem{shao2022spatial}
Yihao Shao, Jianjun Liu, Jinlong Yang, and Zebin Wu.
\newblock Spatial--spectral involution mlp network for hyperspectral image classification.
\newblock {\em IEEE Journal of Selected Topics in Applied Earth Observations and Remote Sensing}, 15:9293--9310, 2022.

\bibitem{suarez2021tutorial}
Juan~Luis Su{\'a}rez, Salvador Garc{\'\i}a, and Francisco Herrera.
\newblock A tutorial on distance metric learning: Mathematical foundations, algorithms, experimental analysis, prospects and challenges.
\newblock {\em Neurocomputing}, 425:300--322, 2021.

\bibitem{wang2019multi}
Xun Wang, Xintong Han, Weilin Huang, Dengke Dong, and Matthew~R Scott.
\newblock Multi-similarity loss with general pair weighting for deep metric learning.
\newblock In {\em Proceedings of the IEEE/CVF conference on computer vision and pattern recognition}, pages 5022--5030, 2019.

\bibitem{wohlwend2019metric}
Jeremy Wohlwend, Ethan~R Elenberg, Samuel Altschul, Shawn Henry, and Tao Lei.
\newblock Metric learning for dynamic text classification.
\newblock {\em arXiv preprint arXiv:1911.01026}, 2019.

\bibitem{zhang2021automatic}
Hao Zhang, Lu Yuan, Guangyu Wu, Fuhui Zhou, and Qihui Wu.
\newblock Automatic modulation classification using involution enabled residual networks.
\newblock {\em IEEE Wireless Communications Letters}, 10(11):2417--2420, 2021.

\bibitem{JMLR:v25:23-0912}
Shijun Zhang, Jianfeng Lu, and Hongkai Zhao.
\newblock Deep network approximation: Beyond relu to diverse activation functions.
\newblock {\em Journal of Machine Learning Research}, 25(35):1--39, 2024.

\bibitem{zhong2021deep}
Aoxiao Zhong, Xiang Li, Dufan Wu, Hui Ren, Kyungsang Kim, Younggon Kim, Varun Buch, Nir Neumark, Bernardo Bizzo, Won~Young Tak, et~al.
\newblock Deep metric learning-based image retrieval system for chest radiograph and its clinical applications in covid-19.
\newblock {\em Medical Image Analysis}, 70:101993, 2021.

\end{thebibliography}
}

\end{document}